\pdfoutput=1

\documentclass[11pt]{article}

\usepackage[]{acl}

\usepackage{times}
\usepackage{latexsym}

\usepackage[T1]{fontenc}

\usepackage[utf8]{inputenc}

\usepackage{enumitem}
\usepackage{microtype}
\usepackage{graphicx} 
\graphicspath{{Figures/}}
\usepackage{amsmath}
\usepackage{amssymb}
\usepackage{bm}
\usepackage{threeparttable}
\usepackage{booktabs}
\usepackage{multicol}
\usepackage{multirow}
\usepackage{booktabs}
\usepackage{array}

\title{TSAM: A Two-Stream Attention Model for Causal Emotion Entailment}
\author{Duzhen Zhang\thanks{\ \ This work was done when Duzhen Zhang was interning at Pattern Recognition Center, WeChat AI, Tencent Inc, China.}, Zhen Yang, Fandong Meng, Xiuyi Chen\thanks{\ \ Corresponding author.}, Jie Zhou\\
        Pattern Recognition Center, WeChat AI, Tencent Inc, Beijing, China\\
        \texttt{\{bladedancer957,hugheren.chan\}}@gmail.com\\
        \texttt{\{zieenyang,fandongmeng,withtomzhou\}}@tencent.com
        }

\begin{document}
\maketitle
\begin{abstract}
Causal Emotion Entailment (CEE) aims to discover the potential causes behind an emotion in a conversational utterance. Previous works formalize CEE as independent utterance pair classification problems, with emotion and speaker information neglected.
From a new perspective, this paper considers CEE in a joint framework. We classify multiple utterances synchronously to capture the correlations between utterances in a global view and propose a Two-Stream Attention Model (TSAM) to effectively model the speaker's emotional influences in the conversational history.
Specifically, the TSAM comprises three modules: Emotion Attention Network (EAN), Speaker Attention Network (SAN), and interaction module. The EAN and SAN incorporate emotion and speaker information in parallel, and the subsequent interaction module effectively interchanges relevant information between the EAN and SAN via a mutual BiAffine transformation.
Extensive experimental results demonstrate that our model achieves new State-Of-The-Art (SOTA) performance and outperforms baselines remarkably.
\end{abstract}

\section{Introduction}

\begin{figure}[htbp]
	\centering  
	\includegraphics[width=6cm]{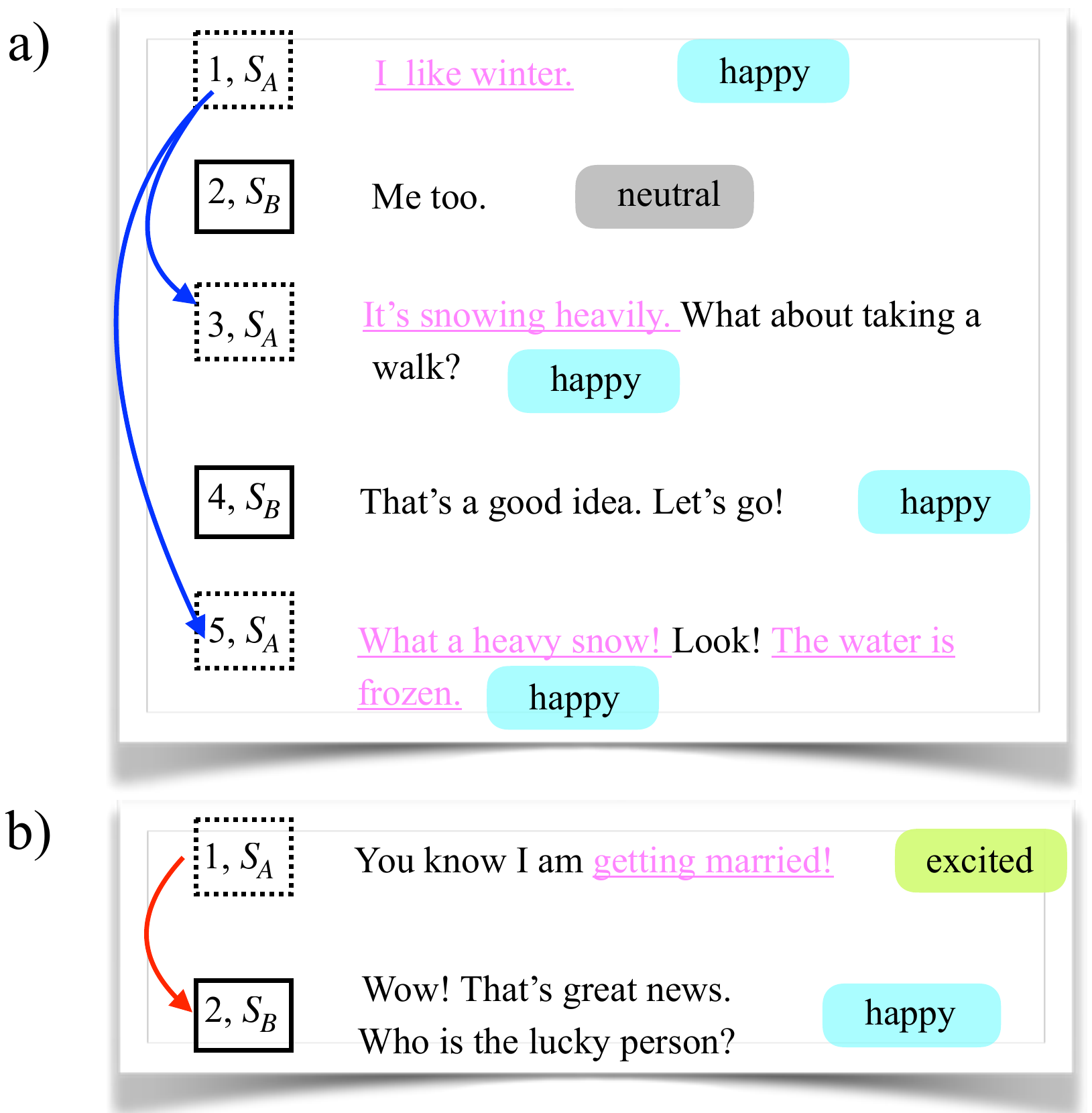}
	\caption{Example conversations sampled from the benchmark dataset~\citep{poria2021recognizing}}\label{fig-1}
\end{figure}
With the recent proliferation of open conversational data on social media platforms, such as Twitter and Facebook, Emotion Analysis in Conversations (EAC) has become a popular research topic in the field of Natural Language Processing (NLP). Most of the existing works on EAC mainly focus on Emotion Recognition in Conversations (ERC), i.e., recognizing emotion labels of utterances (e.g., happy, sad, etc.) ~\citep{poria2017context,poria2019emotion,wang2020contextualized,zhang2020knowledge}. However,~\citet{poria2021recognizing} point out that these studies lack further reasoning about emotions, such as understanding the stimuli and the cause of the emotion. Since Recognizing Emotion Cause in Conversations (RECCON) holds the potential to improve the interpretability and performance of affect-based models,~\citet{poria2021recognizing} put forward a new promising task, named RECCON, which includes two different sub-tasks: Causal Span Extraction (CSE) at word/phrase level and Causal Emotion Entailment (CEE) at utterance level. Due to the simplicity and sufficiency describing emotion causes at the utterance level, we focus on the CEE task in this paper, whose goal is to predict which particular utterances in the conversational history contain the cause of non-neutral emotion in the target utterance. 


Compared to the Emotion Cause Extraction (ECE) in news articles~\citep{gui2016event,xia2019emotion}, CEE is particularly challenging due to the informal expression style and the intermingling dynamic among interlocutors. 
\citet{poria2021recognizing} consider CEE as a set of independent utterance pair classification problems and neglect the emotion and speaker information in the conversational history. Thus, they can neither capture the correlations between contextual utterances in a global view nor model the speaker's emotional influences, namely the intra-speaker and inter-speaker emotional influences.\footnote{The speaker's emotional influences are predominant types of emotion causes in the dataset as shown in Table~\ref{tab-1}.}
Intra-speaker emotional influences mean that the cause of the emotion is primarily due to the speaker's stable mood induced from previous dialogue turns. As shown in Figure~\ref{fig-1} (a), utterance 1 establishes the concept that Speaker A ($S_A$) likes winter, which triggers a happy mood for future utterances 3 and 5. Inter-speaker emotional influences mean that the emotion of the target speaker is induced from an event mentioned or emotion revealed by other speakers. As Figure~\ref{fig-1} (b) shows, $S_B$'s happy emotion may be triggered by the event ``getting married'' mentioned by $S_A$, or by the fact that $S_A$ is excited about getting married. 

To remedy this defect, we tackle CEE in a joint framework. We classify multiple utterances synchronously to capture the correlations between contextual utterances and propose a TSAM to effectively model the speaker's emotional influences in the conversational history. Specifically, the TSAM contains three modules: EAN, SAN, and interaction module. The EAN provides utterance-to-emotion interactions to incorporate emotion information by performing attention over emotion embeddings. The SAN represents different speaker relations between utterances in a graph, which provides utterance-to-utterance interactions to incorporate speaker information by performing attention over the speaker relation graph. These two modules incorporate emotion and speaker information in parallel. Moreover, inspired by~\citep{li2021dual,tang2020dependency}, the interaction module interchanges relevant information between the EAN and SAN through a mutual BiAffine transformation. Finally, the entire TSAM can be stacked in multiple layers to refine iteratively and interchange emotion and speaker information.


\begin{itemize}
\item For the first time, we tackle CEE in a joint framework to capture the correlations between contextual utterances in a global view.
\item We propose a TSAM to model the speaker's emotional influences in the conversational history, which contains EAN, SAN, and interaction module to incorporate and interchange emotion and speaker information.
\item Experimental results on the benchmark dataset~\citep{poria2021recognizing} demonstrate the effectiveness of our proposed model, surpassing the SOTA model significantly.
\end{itemize}

\begin{figure*}[htbp]
	\centering  
	\includegraphics[width=14cm]{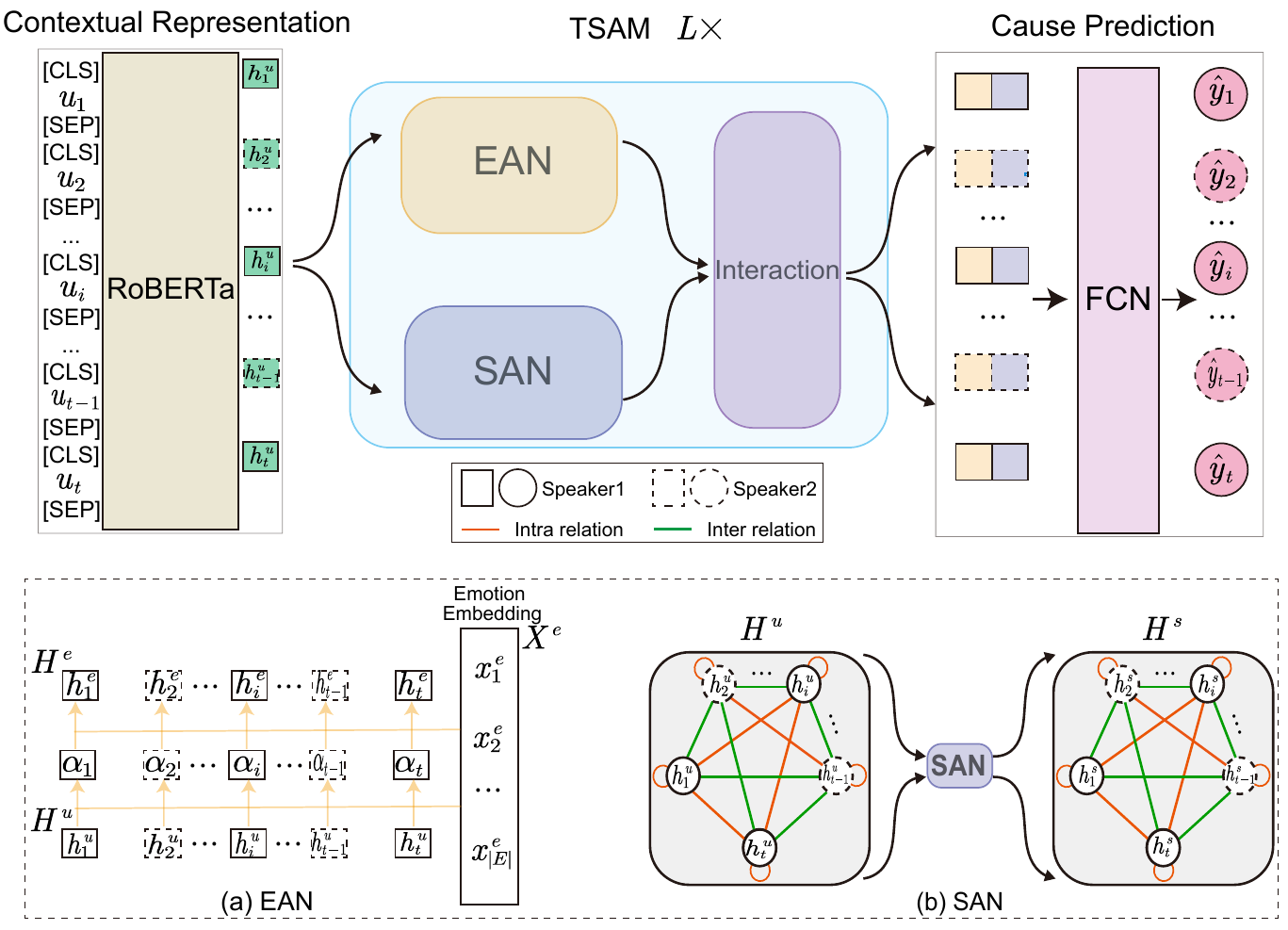}
	\caption{The top is the proposed model's entire architecture, and the bottom is the detailed architecture of model components: (a) EAN, (b) SAN. First, we obtain the contextual representation for each utterance with RoBERTa. Then, the TSAM is utilized to model the speaker's emotional influences in the conversational history. Finally, the cause prediction module is used to output the predictions.
	}\label{fig-2}

\end{figure*}

\section{Related Work}
\paragraph{ECE} Early works mainly exploit rule-based methods~\citep{lee2010text,lee2010emotion,chen2010emotion} to identify the potential causes for certain emotion expressions in the text.~\citet{gui2016event} first release a public annotated dataset for ECE, and based on which some feature based~\citep{gui2016emotion} and neural based methods~\citep{gui2017question,li2018co,ding2019independent,DBLP:conf/ijcai/XiaZD19,DBLP:conf/acl/Yan0PH20,li2021boundary} appear. To extract emotion and its corresponding cause jointly,~\citet{xia2019emotion} first put forward the Emotion-Cause Pair Extraction (ECPE) task and tackle it by a two-step method. Subsequently, many improved methods are proposed to tackle ECPE in an end2end manner~\citep{ding2020ecpe,ding2020end,yuan2020emotion,fan2020transition,wei2020effective,cheng2020symmetric,chen2020unified,chen2020end}. However, these works mentioned above use news articles as the target corpus for ECE, which largely reduces reasoning complexity. By contrast, 
CEE is more challenging due to the intermingling dynamic among interlocutors and the informal expression style.

\paragraph{ERC} Recently, due to the proliferation of publicly available conversational datasets~\citep{zhou2018emotional,chen2019emotionlines,poria2019meld,chatterjee2019understanding,yuqiang2022}, there is a growing number of studies on ERC~\citep{hazarika2018icon,hazarika2018conversational,majumder2019dialoguernn,zhong2019knowledge,jiao2019higru,ghosal2020dialoguegcn,ishiwatari2020relation,ghosal2020cosmic,DBLP:conf/acl/ShenWYQ20,DBLP:conf/acl/ZhuP0ZH20,DBLP:conf/acl/HuWH20,guibon2021few,DBLP:conf/ijcai/ZhaoZL22,peng2022you}.
Although substantial progress has been made in ERC, these studies lack further reasoning about emotions, such as understanding the stimuli or the cause of an emotion expressed by a speaker~\citep{poria2021recognizing}.

\paragraph{RECCON} For further reasoning about emotions,~\citet{poria2021recognizing} propose a new task named RECCON, which contains two different sub-tasks: CSE at word/phrase level and CEE at utterance level.~\citet{poria2021recognizing} formalize CEE as a set of independent utterance pair classification problems, neglecting the emotion and speaker information in the conversational history. Specifically, they pair a target utterance with each utterance in its conversational history and determine whether the utterance contains the cause of emotion in the target utterance. Thus, they cannot capture the correlations between contextual utterances in a global view and fail to model the speaker's emotional influences in the conversational history. From a new perspective, we tackle CEE in a joint framework. We encode and classify multiple utterances synchronously to capture the correlations in a global view and propose a TSAM to model the speaker's emotional influences effectively.

\section{Task Definition}
We first define the task of CEE formally. For a target utterance $u_t$, i.e., the $t^{th}$ utterance in the conversation, the goal of CEE is to predict which particular utterances in the conversational history $L(u_t)=(u_1,u_2,...,u_t)$ are responsible for the non-neutral emotion in the target utterance. $u_i$ is set as a positive example if it contains the cause of non-neutral emotion in the target utterance and a negative example otherwise, where $i=1,...,t$.
The independent utterance pair classification framework~\citep{poria2021recognizing} performs $t$ independent classifications, each of which takes $(u_t, u_i)$ as input. Therefore, it fails to capture the correlations between contextual utterances in a global view. 
On the contrary, the proposed joint classification framework only performs one joint classification with $L(u_t)$ as input,
which makes it possible to capture the correlations between contextual utterances.

\section{Method}
The proposed model consists of three components: the contextual utterance representation, the TSAM, and the cause prediction modules. The whole architecture of our model is illustrated in Figure~\ref{fig-2}.

\subsection{Contextual Utterance Representation}
The pre-trained RoBERTa is utilized as the utterance encoder, and we extract the contextual utterance representations by feeding the whole of the conversational history $L(u_t)$ into the RoBERTa \citep{liu2019roberta}.
Specifically, each utterance in $L(u_t)$ is expanded to start with the token ``[CLS]'' and end with the token ``[SEP]''. The input representation for each token is the sum of its corresponding token and position embeddings. The contextual representation $\bm{h}^u_i \in \mathbb{R}^{d_h}$ for utterance $u_i$ is the output of the corresponding ``[CLS]'' token, where $d_h$ denotes the dimension of the utterance representation. The contextual representation for all utterances is denoted as $\bm{H}^{u} \in \mathbb{R}^{t\times d_h}$. The RoBERTa we utilized is fine-tuned with the training process.

\subsection{TSAM}
The TSAM models the speaker's emotional influences with three modules: EAN, SAN, and Interaction. We first illustrate the calculation process of each module in one-layer TSAM and then generalize it to multiple successive layers.

\subsubsection{EAN}
The EAN provides utterance-to-emotion interactions to explicitly incorporate emotion information by performing attention over emotion embeddings.

\paragraph{Emotion Representation} Given the set of candidate emotion labels $\mathcal{E}=\{e_1,...,e_{|\mathcal{E}|}\}$, each emotion label $e_k$ is represented using an embedding vector \citep{cui2019hierarchically}:
\begin{equation}
       \bm{x}^e_k=\text{Embed}(e_k)\in \mathbb{R}^{d_h}
\end{equation}
where $k=1,...,|\mathcal{E}|$, $d_h$ denotes the dimension of the emotion embedding. $\text{Embed}$ represents an emotion embedding lookup table, which is initialized by contextual embeddings from RoBERTa and tuned during model training. The embedding for the set of the whole emotion labels is denoted as $\bm{X}^e \in \mathbb{R}^{|\mathcal{E}|\times d_h}$. 

\paragraph{EAN Inference} With the emotion labels represented as embeddings, we extract the emotion information $\bm{H}^e\in \mathbb{R}^{t\times d_h}$ by performing dot-product attention over contextual utterance representations and emotion embeddings, which is calculated as:
\begin{align}
    \bm{H}^e &= \textit{attention}(\bm{Q},\bm{K},\bm{V}) = \bm{\alpha}\bm{V}\\
    \bm{\alpha} &= \textit{softmax}(\frac{\bm{Q}\bm{K}^T}{\sqrt{d_h}})
\end{align}
where $\bm{Q}=\bm{H}^u, \bm{K}=\bm{V}=\bm{X}^e$, $\bm{\alpha}\in\mathbb{R}^{t\times |\mathcal{E}|}$ is an attention matrix consisting of potential emotion distributions for all utterances. Compared to the standard attention mechanism above, it may be beneficial to use multi-head attention \citep{vaswani2017attention} to capture multiple emotion distributions in parallel and obtain richer emotion information:
\begin{align}
    \bm{H}^e &= \textit{concat}(head_1,...,head_m)\\
    head_j &= \textit{attention}(\bm{Q}\bm{W}^{Q}_j,\bm{K}\bm{W}^K_j,\bm{V}\bm{W}^V_j)
\end{align}
where $\bm{W}^Q_j,\bm{W}^K_j,\bm{W}^V_j\in\mathbb{R}^{d_h\times\frac{d_h}{m}}$ are learnable parameters and $m$ is the number of parallel heads.

Since the emotion labels of the utterances in the conversational history are known, we can also simply use the embedding of emotion label corresponding to the utterance as the extracted emotion information:
\begin{equation}
     \bm{H}^e = \Tilde{\bm{X}}^{e}
\end{equation}
where $\Tilde{\bm{X}}^{e}\in\mathbb{R}^{t\times d_h}$ is the embedding of emotion labels corresponding to all utterances in the history. We refer to the method as Direct Application Emotional Embedding (DAEE). Compared with DAEE, the potential advantages of the EAN are as follows: (1) The EAN can provide utterance-to-emotion interactions and capture multiple potential emotion distributions through multi-head attention to obtain more comprehensive and richer emotion information; (2) The soft emotion distributions can model the mutual impact among different emotions for further enhancement of emotion embeddings, while each emotion embedding is relatively independent of each other in DAEE; (3) The EAN can avoid emotion annotation errors to a certain extent. We apply EAN in our model to incorporate emotion information by default and compare the EAN and DAEE in the part of experiments.

\subsubsection{SAN}
The SAN provides utterance-to-utterance interactions to incorporate speaker information by performing attention over the speaker relation graph.

\paragraph{Graphical Structure} We define a conversational history with $t$ utterances as a graph $\mathcal{G}=(\mathcal{V},\mathcal{E},\mathcal{R})$, with nodes (utterances) $v_i\in \mathcal{V}$ and labeled edges (relations) $(v_i,r,v_j)\in \mathcal{E}$, where $r\in \mathcal{R}$ is a relation type. We also add a self-loop edge to every node, as the cause may be present within the target utterance itself. The representation of node $v_i$ is initialized with the contextual utterance representation $\bm{h}^u_i\in \mathbb{R}^{d_h}$, i.e., the $i^{th}$ embedding in $\bm{H}^u$. There are two relation types of edges: (1) \textbf{Intra} relation type: how the utterance influences other utterances (including itself) expressed by the same speaker; (2) \textbf{Inter} relation type: how the utterance influences ones expressed by other speakers. 

\paragraph{SAN Inference} The representation of a node $\bm{h}_i$ is updated by aggregating representations of its neighborhood $\mathcal{N}^r(i)$ under the relation type $r$. The graph attention mechanism \citep{velivckovic2018graph} is used to attend to the neighborhood's representations. The output of a node $\bm{h}^s_i\in \mathbb{R}^{d_h}$ is calculated as the sum of the hidden features $\bm{h}_{ir}\in \mathbb{R}^{d_h}$ under relation $r$. The propagation is defined as follows:
\begin{align}
    \alpha_{ijr} &=\textit{softmax}_i(\textit{LRL}(\bm{a}^T_r[\bm{W}_r\bm{h}^u_i;\bm{W}_r\bm{h}^u_j]))\\
    \bm{h}_{ir} &= \sum_{j\in\mathcal{N}^r(i)}\alpha_{ijr}\bm{W}_r\bm{h}^u_j\label{equ}\\
    \bm{h}^s_i &= \sum_{r\in \mathcal{R}}\bm{h}_{ir}
\end{align}
where $\alpha_{ijr}$ denotes the edge weight from utterance $u_i$ to its neighborhood $u_j$ under relation type $r$, $\bm{W}_r\in \mathbb{R}^{d_h\times d_h}$ and $\bm{a}_r\in\mathbb{R}^{d_h}$ denote a learnable weight matrix and a vector for each relation type $r$ respectively. $\textit{LRL}$ denotes $\textit{LeakyReLU}$ activation function. The updated representation of all nodes is denoted as $\bm{H}^{s} \in \mathbb{R}^{t\times d_h}$.

\subsubsection{Interaction Module}
To effectively interchange relevant information between the EAN and SAN, we apply a mutual BiAffine transformation as a bridge. The calculation process is formulated as follows:
\begin{align}
    \bm{A}_1 &= \textit{softmax}(\bm{H}^e\bm{W}_1(\bm{H}^s)^T)\\
    \bm{A}_2 &= \textit{softmax}(\bm{H}^s\bm{W}_2(\bm{H}^e)^T)\\
    \bm{H}^{e'} &= \bm{A}_1\bm{H}^s\\
    \bm{H}^{s'} &= \bm{A}_2\bm{H}^e
\end{align}
where $\bm{W}_1,\bm{W}_2\in \mathbb{R}^{d_h\times d_h}$ are trainable parameters and $\bm{A}_1, \bm{A}2\in \mathbb{R}^{t\times t}$ are temporary alignment matrices projecting from $\bm{H}^s$ to $\bm{H}^e$ and $\bm{H}^e$ to $\bm{H}^s$, respectively. Here, $\bm{H}^{e'}\in \mathbb{R}^{t\times d_h}$ can be viewed as a projection from $\bm{H}^s$ to $\bm{H}^e$ , and $\bm{H}^{s'}\in \mathbb{R}^{t\times d_h}$ follows the same principle.

\subsubsection{The Whole Process}
We generalize the TSAM to multiple successive layers to iteratively refine and interchange emotion and speaker information. The detailed procedures are as follows:
\begin{align}
   \bm{H}^e_l &= \textbf{EAN}(\bm{E}_l, \bm{X}^e) \\
   \bm{H}^s_l &= \textbf{SAN}(\bm{S}_l) \\
   \bm{H}^{e'}_l, \bm{H}^{s'}_l &= \textbf{Interaction}(\bm{H}^e_l,  \bm{H}^s_l)\\
   \bm{E}_{l+1}, \bm{S}_{l+1} &=  \bm{H}^{e'}_l, \bm{H}^{s'}_l
\end{align}
where $\bm{E}_0=\bm{S}_0=\bm{H}^u$. The TSAM can be stacked in $L$ layers and $l\in [0,L-1]$.

\subsection{Cause Prediction}
We obtain the final utterance representation for $u_i$ by concatenating the output $(\bm{E}_L,\bm{S}_L)$ of the L-layer TSAM. Finally, the concatenated vector is classified using a Fully-Connected Network (FCN):
\begin{align}
    \bm{l}_i &= \textit{ReLU}(\bm{W}_1[\bm{e}^L_i;\bm{s}^L_i]+\bm{b}_1)\\
    \hat{y}_i &= \textit{sigmoid}(\bm{W}_2\bm{l}_i+b_2)
\end{align}
where $\hat{y}_i$ is the probability for utterance $u_i$ containing the cause of emotion in the target utterance, $\bm{e}^L_i, \bm{s}^L_i\in \mathbb{R}^{d_h}$ denote the $i^{th}$ embedding in $\bm{E}_L$ and $\bm{S}_L$, respectively, $\bm{W}_1\in\mathbb{R}^{d_h\times 2d_h}, \bm{W}_2\in \mathbb{R}^{1\times d_h}, \bm{b}_1\in \mathbb{R}^{d_h}$ and $b_2$ are learnable parameters of FCN.

\section{Experimental Settings}
\subsection{Dataset and Evaluation Metrics}

\begin{table}[tbp]
\centering
\fontsize{10}{13}\selectfont 
\renewcommand\tabcolsep{3.0pt}
\begin{tabular}{c|cc|c}
\hline
\multicolumn{3}{c|}{Statistics}                                                                                               & RECCON-DD \\ \hline\hline
\multirow{6}{*}{\begin{tabular}[c]{@{}c@{}}Data\\ Distributions\end{tabular}}             & \multirow{2}{*}{Train} & Positive & 7269  \\
                                                                                          &                        & Negative & 20646  \\ \cline{2-4}
                                                                                          & \multirow{2}{*}{Dev}   & Positive & 347  \\
                                                                                          &                        & Negative & 838  \\ \cline{2-4}
                                                                                          & \multirow{2}{*}{Test}  & Positive & 1894  \\
                                                                                          &                        & Negative & 5330  \\ \hline
\multirow{5}{*}{\begin{tabular}[c]{@{}c@{}}Cause\\ Type\\ Distributions\end{tabular}}     & \multicolumn{2}{c|}{No Context}   & 43\%  \\
                                                                                          & \multicolumn{2}{c|}{Inter}        & 32\%  \\
                                                                                          & \multicolumn{2}{c|}{Intra}        & 9\%  \\
                                                                                          & \multicolumn{2}{c|}{Hybrid}       & 11\%  \\
                                                                                          & \multicolumn{2}{c|}{Unmentioned}          & 5\%  \\ \hline
\end{tabular}

\caption{Statistics of the RECCON-DD dataset. \textit{No Context}: The cause is present within the target utterance itself; \textit{Inter}: Inter-speaker emotional influences; \textit{Intra}: Intra-speaker emotion influences (Self-Contagion); \textit{Hybrid}: Inter and Intra can jointly cause the emotion of an utterance; \textit{Unmentioned}: Some instances have no explicit emotion causes in the conversational history.}
\label{tab-1}
\end{table}
	
We evaluate the proposed model on a benchmark dataset for RECCON, named RECCON-DD~\citep{poria2021recognizing}, which is constructed based on DailyDialog dataset \cite{li2017dailydialog}.\footnote{DailyDialog is a natural human communication dataset which is usually used in ERC task. It contains utterance-level emotion labels and covers various topics related to daily lives.} Some statistics about RECCON-DD are reported in Table~\ref{tab-1}. Following~\citep{poria2021recognizing}, the macro-averaged F1 score is utilized as the evaluation metric in this paper. We also report the F1 score for both positive and negative samples, denoted as Pos. F1 and Neg. F1 respectively.

\begin{table*}[tbp]
\centering
\begin{tabular}{@{}c|c|ccccccc@{}}
\hline
\multirow{2}{*}{\#} & \multirow{2}{*}{Model} & \multicolumn{3}{c}{W/O CH}   &  & \multicolumn{3}{c}{W/ CH}  \\ \cmidrule(lr){3-5} \cmidrule(l){7-9} 
                    &                        & Pos. F1 & Neg. F1 & macro F1 &  & Pos.F1 & Neg.F1 & macro F1 \\ \hline\hline
0                   & $\text{INDEP}_{base}$            &    56.64     &    85.13     &   70.88       &  &    64.28    &    88.74    &   76.51       \\
1                   & $\text{INDEP}_{large}$           &    50.48     &    87.35     &    68.91      &  &   66.23     &     87.89   &     77.06     \\ \hline
2                   & $\text{JOINT}_{base}$             & -       & -       & -        &  &   66.61     &    89.11    &   77.86       \\
3                   & $\text{JOINT}_{large}$           & -       & -       & -        &  &   68.30     &  89.16      &      78.73    \\ \hline
4                   & $\text{Ours}_{base}$              & -       & -       & -        &  &     68.59   &     89.75   &    79.17      \\
5                   & $\text{Ours}_{large}$           & -       & -       & -        &  &  $\textbf{70.00}^{\dagger}$       &    $\textbf{90.48}^{\dagger}$    &    $\textbf{80.24}^{\dagger}$      \\ \hline
\end{tabular}
\caption{Performance of our model and baselines on the test set of RECCON-DD. Bold font denotes the best performance.
``Ours'' denotes the proposed model without removing any module (``Ours'' = ``JOINT'' + TSAM). ``$^{\dagger}$'' denotes that $\text{Ours}_{large}$ is statistically significant~\citep{koehn2004statistical} better than $\text{INDEP}_{large}$ W/ CH ($p$-$\text{value} < 0.05$).}
\label{tab-2}
\end{table*}

\subsection{Baselines}
For a comprehensive performance evaluation, we compare our model with the following baselines:  

\paragraph{(1) $\text{INDEP}_{base}$}\citep{poria2021recognizing} It tackles CEE in an independent classification framework (INDEP) and uses the RoBERTa-Base model~\citep{liu2019roberta} as the utterance encoder. The input is formated as "[CLS]$u_t$[SEP]$u_i$[SEP]" and the classification is performed from the final representation of the token "[CLS]". 
\paragraph{(2) $\text{INDEP}_{large}$}\citep{poria2021recognizing} Compared to (1), it uses the RoBERTa-Large model as utterance encoder; 
\paragraph{(3) $\text{JOINT}_{base}$} It's one of the variants of our model, where the TSAM is removed. It tackles RECCON in a joint classification framework (JOINT) and uses the RoBERTa-Base model as the utterance encoder. Moreover, its input format is "[CLS]$u_1$[SEP][CLS]$u2$[SEP],...,[CLS]$u_t$[SEP]" and the classifications are performed synchronously from the corresponding contextual utterance representations of the [CLS] tokens; 
\paragraph{(4) $\text{JOINT}_{large}$} Compared to (3), it uses RoBERTa-Large model as the utterance encoder.

For \textbf{INDEP} baselines, there are two different settings: With Conversational History (W/ CH) and Without Conversational History (W/O CH). W/ CH means considering the conversational history. When performing utterance pair classification, the conversational history $L(u_t)$ is concatenated after the input to incorporate contextual information, while W/O CH means ignoring the history.

\subsection{Implementation Details}
Our model's base and large versions use pre-trained RoBERTa-Base and RoBERTa-Large models as the utterance encoders, respectively.\footnote{Our RoBERTa models are adapted from this implementation: \url{https://github.com/huggingface/transformers}} The binary cross-entropy loss along with L2-regularization is used during training, where the coefficient of L2 term is $0.01$ in the RoBERTa structure and $1e$-$5$ in other structures. We set the dropout rate to $0.1$. The learning rate and the batch size are set as $1e$-$5$ and $2$, respectively. Our model is trained with the Adam optimizer~\citep{kingma2015adam}. We set the dimensions of the contextual utterance representation $d_h$ as 768/1024 in the Base/Large version of the proposed model. We use $4$-$ $head attention in EAN, and the number of TSAM layers $L$ is set to $3$. We train the model for $40$ epochs in total and use the early stopping strategy based on the performance on the development set. Then, the model with the highest macro-averaged F1 score is used to evaluate the test set. Other hyper-parameters are selected according to the performance of the development set. All of the experiments are conducted on an NVIDIA V100 GPU with $32$GB of memory.

\section{Results and Discussions}

\subsection{Main Results}
Experimental results are reported in Table~\ref{tab-2}. We directly cite the results for the baselines reported in ~\citep{poria2021recognizing}. For the performance of each model we implemented, we report the average score of 5 runs. From Table~\ref{tab-2}, we can find that the proposed model (\#5) outperforms all of the baselines and surpasses the best model (\#1, W/ CH) in~\citep{poria2021recognizing} with more than 3 points of macro F1 score.

Further comparisons show that models with the large pre-trained utterance encoder are more likely to achieve better performance (about 1 point of macro F1 score) than the corresponding models with the base one, except for the models under W/O CH setting in the Table~\ref{tab-2}. 
By comparing two different settings W/O CH and W/ CH in Table~\ref{tab-2}, we can find that the conversational history plays a significant role for $\text{INDEP}_{base}$ and $\text{INDEP}_{large}$ models. This is mainly because that the conversational history is able to provide the contextual information for prediction. 
Due to the simultaneous classification of multiple utterances in the conversational history under the joint framework, $\text{JOINT}_{base}$ and $\text{JOINT}_{large}$ models can naturally incorporate the contextual information. 
The $\text{JOINT}_{base}$ and $\text{JOINT}_{large}$ models significantly outperform the $\text{INDEP}_{base}$ W/ CH and $\text{INDEP}_{large}$ W/ CH models by about 1.5 points of macro F1 scores respectively (comparing \#0 with \#2, and \#1 with \#3 in Table 2). There may be two main factors: 1) Simply concatenating the conversational history after the utterance pair to be classified in INDEP W/ CH models may destroy the structure of the conversation; 2) Compared to INDEP W/ CH models, classification of multiple utterances synchronously in JOINT models will have more sufficient supervision signals and can more effectively model the correlations between contextual utterances in a global view, i.e., utterances with similar semantics are supposed to have similar chances being the emotion cause.
The comparison between \#2 and \#4 (or \#3 and \#5) in Table~\ref{tab-2} shows the effectiveness of the proposed TSAM. The model with TSAM (\#5) achieves an improvement up to 1.51 points of macro F1 score than the model without TSAM (\#3).

\subsection{Ablation Study}
In this subsection, we conduct ablation studies to analyze the effects of different components based on $\text{Ours}_{large}$ mentioned in Table~\ref{tab-2}.

\paragraph{Effect of Emotion Information}

We compare three different ways for incorporating the emotion information: no emotion information incorporated, incorporating emotion information with direct application emotional embedding, and incorporating emotion information with EAN. The results are shown in Table~\ref{tab-3}.
We can find that the performance of the proposed model degrades if the emotion information is not incorporated (comparing row 1 with 3 in Table~\ref{tab-3}). This result shows that the emotion information in the conversational history plays a significant role in the task of CEE. By comparing rows 2 with 3 in Table~\ref{tab-3}, the result shows that EAN achieves better performance than DAEE since EAN can extract richer emotion information and model the mutual impact among different emotions.

\begin{table}[tbp]
\centering
\resizebox{0.98\hsize}{!}{%
\begin{tabular}{c|ccc}
\hline
Emotion Information  & Pos. F1 & Neg. F1&macro F1 \\ \hline\hline
No                            & 68.40   & 89.80 & 79.10     \\
DAEE       &68.90  &  90.03	  &   79.47  \\
      EAN                      & \textbf{70.00} & \textbf{90.48} &   \textbf{80.24}    \\ \hline
\end{tabular}}
\caption{Comparison of different ways of incorporating emotion information. \textit{No}: no emotion information incorporated; \textit{DAEE}: incorporating the emotion information with direct application emotional embedding.}
\label{tab-3}
\end{table}

\paragraph{Effect of Speaker Information}
To evaluate the effects of speaker information, we remove the speaker relations in SAN, resulting in a single edge relation throughout the graph. As Table~\ref{tab-4} shows, the performance of our model decreases dramatically if not considering the speaker information. This result presents that modeling the speaker information in the conversational history is very important for the final performance.

\begin{table}[htbp]
\centering
\resizebox{0.98\hsize}{!}{%
\renewcommand\tabcolsep{3.0pt}
\begin{tabular}{c|ccc}
\hline
Speaker Information & Pos. F1 & Neg. F1&macro F1 \\ \hline\hline
Not Consider        & 67.99 &   89.42 &     78.71     \\
Consider        & \textbf{70.00} & \textbf{90.48} &   \textbf{80.24}       \\ \hline
\end{tabular}}
\caption{Results on experiments whether considering speaker information or not in SAN.}
\label{tab-4}
\end{table}

\paragraph{Effect of Interaction Module}
We remove the interaction module in each layer so that the EAN and SAN can't interact. As Table~\ref{tab-5} shows, the performance of our model decreases dramatically when the interaction module is removed. This result shows that the effective interchange of relevant information between EAN and SAN is conducive to the final performance.
\begin{table}[htbp]
\centering
\resizebox{0.98\hsize}{!}{%
\renewcommand\tabcolsep{3.0pt}
\begin{tabular}{c|ccc}
\hline
 & Pos. F1 & Neg. F1&macro F1 \\ \hline\hline
W/O Interaction         & 68.18 &   88.93 &     78.56     \\
W/ Interaction         & \textbf{70.00} & \textbf{90.48} &   \textbf{80.24}       \\ \hline
\end{tabular}}
\caption{Results on experiments whether removing interaction module or not in TSAM.}
\label{tab-5}
\end{table}

\paragraph{Ability on Modeling Emotional Influences}
To evaluate the proposed model's ability to model the speaker's emotional influences, we collect the positive examples from the test set where the causes are induced from the inter-speaker or intra-speaker emotional influences. And we test the prediction accuracy on the collected samples for the proposed $\text{Ours}_{large}$ with and without TSAM. As shown in Table~\ref{tab-6},  W/ TSAM outperforms W/O TSAM by around $2$ points on both cause types, which further verifies that the TSAM can effectively model the emotional influences between speakers.
\begin{table}[tbp]
\centering
\renewcommand\tabcolsep{8.0pt}
\begin{tabular}{c|cc}
\hline
Models & Intra & Inter \\ \hline\hline
W/O TSAM   & 62.06 & 72.67      \\
W/ TSAM  & \textbf{63.82} & \textbf{74.81}  \\ \hline
\end{tabular}
\caption{Accuracy on the collected samples. \textit{Intra}: Intra-speaker emotional influences;  \textit{Inter}: Inter-speaker emotional influences.}
\label{tab-6}
\end{table}

\subsection{Impact of the TSAM Layer Number}
\begin{figure}[htbp]
	\centering  
	\includegraphics[width=7.5cm,height=5cm]{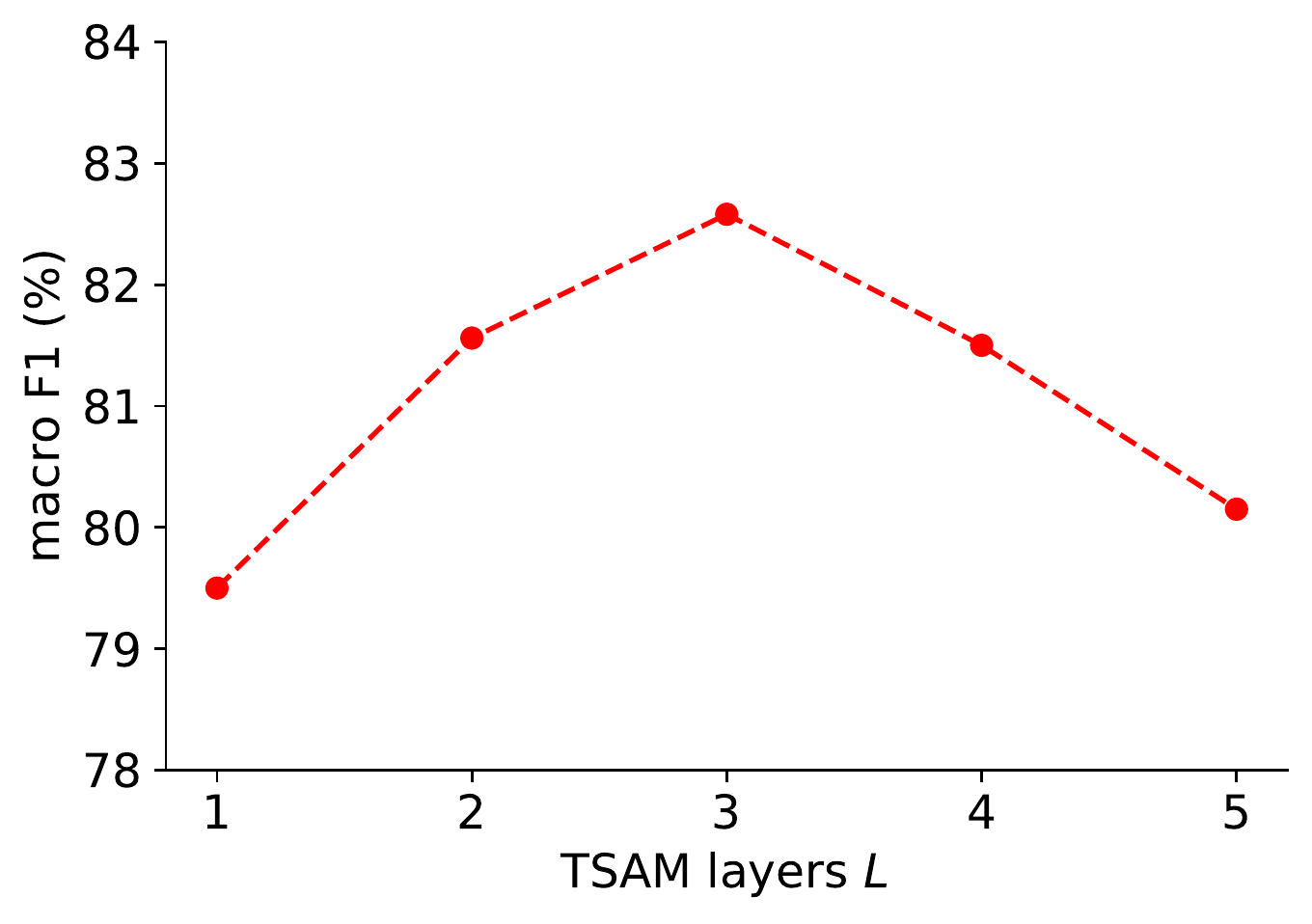}
	\caption{Results of $\text{Ours}_{large}$ with various TSAM layers on the development set of RECCON-DD.}\label{fig-3}
\end{figure}

Since TSAM for modeling speakers' emotional influences is the critical component of our model, we chose the number of TSAM layers $L$ (ranging from 1 to 5) on the development set of RECCON-DD. As shown in Figure~\ref{fig-3}, our model with three TSAM layers achieves the best performance. On the one hand, emotion and speaker information may not be refined and interchanged well when the number of layers is small. On the other hand, if the number of layers is excessive, the performance will decrease, possibly due to information redundancy.

\subsection{Error Analysis}
By analyzing our predicted emotion causes, we find that the following aspects mainly cause the predicted errors. Firstly, our model weakly gives the correct predictions for target utterances with three or more causes.\footnote{Utterances with 3 or more causes account for approximately 14\% of the RECCON-DD dataset}  Compared to the utterances with 1 or 2 causes, the proposed model dropped six macro F1 scores on utterances with multiple causes. Secondly, our model cannot predict well when the underlying emotional cause is latent. At this point, recognizing emotion causes may require complex reasoning steps, and commonsense knowledge is an integral part of this process. We take the case below as an example:
\begin{itemize}
\item $S_A$ (happy): Hello, thanks for calling 123 Tech Help. I'm Todd. How can I help you?

\item $S_B$ (fear): Hello? Can you help me? My computer! Oh, man...
\end{itemize}
In this case, $S_A$ is happy to help $S_B$. In this example, the cause of happy emotion is due to the event ``greeting'' or intention to provide help. On the other hand, $S_B$ is fearful because his or her computer is broken. Both speakers' causes of elicited emotions can only be inferred using commonsense knowledge, which our model does not explicitly consider.

\section{Conclusion and Future Work}
For the first time, we tackle CEE in a joint framework. We classify multiple utterances synchronously to capture the correlations between contextual utterances in a global view and propose a TSAM to effectively model the speaker's emotional influences.
Experimental results on the benchmark dataset show that our model significantly outperforms the SOTA model, and further analysis verifies the effectiveness of each component. This paper points out a new reliable route for follow-up works: incorporating the emotion and speaker information explicitly and modeling the speaker's emotional influences effectively can bring enormous benefits for the tasks similar to CEE.

In the future, we would explore three aspects: (1) Learn emotion recognition and emotion cause recognition in conversations jointly; (2) Apply our model to other similar tasks which need to incorporate the speaker and emotion information; (3) Incorporate commonsense knowledge into the model explicitly to address situations when the underlying emotion cause is latent.

\bibliography{anthology,custom}
\bibliographystyle{acl_natbib}


\end{document}